\pdfoutput=1

\documentclass[11pt]{article}

\usepackage[]{emnlp2021}

\usepackage{times}
\usepackage{latexsym}

\usepackage[T1]{fontenc}

\usepackage[utf8]{inputenc}

\usepackage{microtype}

\usepackage{amsmath,amsfonts,bm}

\def\eqref#1{equation~\ref{#1}}

\def\floor#1{\lfloor #1 \rfloor}
\def\1{\bm{1}}

\DeclareMathAlphabet{\mathsfit}{\encodingdefault}{\sfdefault}{m}{sl}
\SetMathAlphabet{\mathsfit}{bold}{\encodingdefault}{\sfdefault}{bx}{n}

\newcommand{\E}{\mathbb{E}}

\usepackage{hyperref}
\usepackage{url}
\usepackage{fixltx2e}
\usepackage{amsmath}
\usepackage{bm}
\usepackage{color}
\usepackage{graphicx}
\usepackage{subfig}
\usepackage{breqn}
\usepackage{algorithmic}
\usepackage[ruled,vlined]{algorithm2e}

\title{Efficient Reinforcement Learning for Unsupervised Controlled Text Generation}

\author{Bhargav Upadhyay \\
  \texttt{bhargav@agaralabs.com} \\
  \AND
  Akhilesh Sudhakar \\
  \texttt{akhilesh@agaralabs.com} \\\And
  Arjun Maheswaran \\
  \texttt{arjun@agaralabs.com} \\}

\begin{document}
\maketitle
\begin{abstract}
Controlled text generation tasks such as unsupervised text style transfer have increasingly adopted the use of Reinforcement Learning (RL). A major challenge in applying RL to such tasks is the sparse reward, which is available only after the full text is generated. Sparse rewards, combined with a large action space make RL training sample-inefficient and difficult to converge. Recently proposed reward-shaping strategies to address this issue have shown only negligible gains. In contrast, this work proposes a novel approach that provides dense rewards to each generated token. We evaluate our approach by its usage in unsupervised text style transfer. Averaged across datasets, our style transfer system improves upon current state-of-art systems by 21\% on human evaluation and 12\% on automatic evaluation. Upon ablated comparison with the current reward shaping approach (the `roll-out strategy'), using dense rewards improves the overall style transfer quality by 22\% based on human evaluation. Further the RL training is 2.5 times as sample efficient, and 7 times faster.

\end{abstract}

\section{Introduction}
Reinforcement Learning (RL) has been used in NLG tasks including neural machine translation (NMT) ~\citep{Bahdanau2017AnAA}, abstractive summarization ~\citep{Paulus2018ADR}, and specifically in controlled text generation tasks such as paraphrase generation ~\citep{li-etal-2018-paraphrase}, sentence simplification ~\citep{zhang-lapata-2017-sentence} and unsupervised text style transfer~\citep{ijcai2019-711, gong2019reinforcement}. 

Unsupervised text style transfer is the task of re-writing text of a given style into a target style without using a parallel corpus of source style and target style sentences for training. For example, in the case of sentiment transfer~\footnote{Similar to previous works, we use a broad socio-linguistic definition of style that includes sentiment and formality}, \textit{`the pizza was rather bland'} can be re-written with positive sentiment as \textit{`the pizza was quite delicious'}. 

RL has gained popularity in this task as it allows to directly optimize for commonly agreed upon objectives of style transfer, i.e, the generated sentences should 1) possess the target style, 2) preserve the non-stylistic parts (content) of the source sentence, and 3) be fluent and natural sounding. However these objectives are typically measured only at the level of the entire generated text and hence, any RL rewards based on them are only available as ‘sparse rewards’. For example,~\citet{gong2019reinforcement} use a style classifier model to provide a target style reward, which requires the complete generated text. On the other hand, generative RL models generate token-by-token, and require ‘credit assignment’ of reward to each individual token, based on their contribution to the overall sparse reward.

To address this problem, previous work uses different ‘reward shaping’ strategies over the sparse reward, of which, the ‘roll-out strategy’ based on Monte Carlo Tree Search~\citep{mcts} is the commonly used approach. In the context of style transfer,~\citet{gong2019reinforcement} use the roll-out strategy while~\citet{ijcai2019-711} use a naive reward shaping strategy of assigning the same discounted sparse reward to every token. However, as observed by~\citet{Wu2018ASO}, the gains from these reward shaping strategies are negligible in practice. Since NLG tasks have a large action space (the vocabulary), sparse rewards leads to high variance in estimating ‘returns’ for each token used in policy gradient updates. This makes the training sample-inefficient, and can result in a sub-optimal policy. 

In contrast to the roll-out strategy, we propose a novel approach to provide dense rewards directly to each generated token. The primary differentiating factor is that these dense rewards are obtained directly through reward models trained using the same training style transfer data.

Our method of providing dense rewards is inspired by~\citet{Shi2018TowardDT} and~\citet{Sudhakar2019TransformingDR}.~\citet{Shi2018TowardDT} propose using Inverse Reinforcement Learning (IRL)~\citep{ZiebartMBD08} to compute dense rewards for unsupervised text generation. Their objective is to learn un-conditional generation of sentences that belong to a particular distribution for which they use a language model’s token-level likelihood as a dense reward approximator.
~\citet{Sudhakar2019TransformingDR} disentangle style tokens and content tokens using self-attention scores of trained BERT style-classifier~\citep{devlin2018bert}. Style tokens are those that contribute to the style of the overall text. Content tokens are non-style tokens.

We propose separate dense rewards at the token-level for each of target style, content preservation and fluency. a) For the style reward, we propose the use of self-attention scores of a trained style classifier~\citep{Sudhakar2019TransformingDR}. b) For content preservation, we propose a simple n-gram based  dense reward, motivated by BLEU~\citep{papineni-etal-2002-bleu}, and applicable to only content tokens. c) For the fluency reward, we use a pre-trained language model's token-level likelihood, motivated by~\citet{Shi2018TowardDT}.

The main contribution of our work is in that we propose a method that addresses the sparse reward problem for RL-based unsupervised text style transfer. We demonstrate how the resulting dense rewards can be used in designing a state-of-art RL-based style transfer system. This method, as compared to the roll-out strategy is a) sample efficient, b) converges to a better reward and c) computes token level reward in $O(T)$ as against $O(T^{2})$, where $T$ is the number of tokens in the output. The method we propose is extensible to other controlled text generation problems.

\section{Related Work}

The previous section refers to RL-based related work in unsupervised text style transfer. We discuss a few more aspects about these works here.~\citet{xu-etal-2018-unpaired} uses a cycled RL approach where a neutralization model disentangles the content and attributes, and then passes the content to an emotionalization model. Similar to our work,~\citet{gong2019reinforcement} use three different rewards to control the style, content preservation and fluency. Their work also performs an MLE pre-training step prior to the RL training.~\citet{ijcai2019-711} use two RL models - one for each direction of style transfer, provide rewards for style and content, and do not disentangle content and style tokens. They alternate between MLE training and RL training steps.

Non-RL based approaches too have been used for unsupervised text style transfer. Some of these works encode style and content in separate latent representations, and decode the style-dependent output from these representations~\citep{fu2018style,pmlr-v70-hu17e, yang2018unsupervised}. Other works that attempt to learn disentangled latent representations include~\citet{john-etal-2019-disentangled,zhao2018language, wu-etal-2019-hierarchical-reinforced, liu2019revision}. A few others explicitly identify attribute and content words from input texts and then train models to generate target style sentences from the content~\citep{li-etal-2018-delete, Sudhakar2019TransformingDR, wu2019mask}. A few recent works~\citep{lample-etal-2018-phrase, ijcai2019-711} avoid style-content disentanglement altogether. 

RL has also been used for unconditional text generation tasks.~\citet{Shi2018TowardDT} propose the use of Inverse RL (IRL),~\citet{Schmidt2020BERTAA} use a retrieval-based method, comparing the generated sentence with the retrieved reference using contextual BERT embeddings to get compute dense rewards, while ~\citet{yu2017seqgan} use the roll-out strategy.

\section{Problem Statement}
We assume a dataset $D =  \{(x_1, s_1), . . . ,(x_m, s_m)\}$ where each sentence $x_i$ is associated with a specific style $s_i \in S$. For instance, for the task of sentiment transfer, $S = \{$`Positive', 'Negative'$\}$, and for formality transfer, $S = \{$`Formal', `Informal'$\}$. We then aim to learn the conditional distribution $ P(y | x, s^{tgt}) $ such that the target ($tgt$) style of $y$ is $s^{tgt}$, and $y$'s content is similar to that of $x$. `Content' refers to the non-stylized part of the input.

\section{Our Approach}
\label{approach}
We introduce the efficient Dense (Reward) Reinforcement Learning based Style Transformer (hereby, DRL). DRL takes the source sentence $x$ of style $s^{src}$ as input and generates the output sentence ${y}$ of style $s^{tgt}$. More formally, it learns:
\begin{equation} \label{eq:brlst} 
P({y}|x,s^{tgt} ; \theta)
\end{equation}
The RL-based setup consists of: a) a generative model, b) the training strategy and c) the rewards.

\subsection{Model}
The architecture of DRL is a decoder-only Transformer, based on the implementation of the Generative Pre-trained Transformer~\citep{radford2018improving} (hereby, GPT), and is similar to that used by~\citet{Sudhakar2019TransformingDR}. The input to the model is the source sentence, as well as a marker indicating the target style, and the output is a sentence in the target style. We use the same input representation as~\citet{Sudhakar2019TransformingDR}.

Following common practice \citep{ijcai2019-711, Wu2018ASO} the training takes place in two phases: a) MLE pre-training, followed by b) RL training.

\subsection{MLE Pre-training}
\label{sec:mle}
As shown by previous work ~\citep{ijcai2019-711, Wu2018ASO, gong2019reinforcement} supervised or pseudo-supervised training helps with faster and sample efficient convergence. We pre-train DRL on a synthetically generated parallel corpus. We create this parallel corpus by using previous style transfer models, hereby `baseline models' (described in section~\ref{sec:baseline models}) on the original non-parallel corpus. This pre-training is performed by minimizing the standard MLE objective. 

\subsection{RL Training}
DRL optimizes a policy using the policy gradient, to maximize a long term reward $J(\theta)$. The parameters of the model $\theta$ define a policy $\pi_\theta(y_t | s_t)$ which maps a state $s_{t}$ to an action $y_t$. The state $s_t$ in our case in the input sentence $x$ concatenated with the tokens generated by the policy till time $t-1$, i.e., $s_t = (x,y_{1..t-1})$. The action $y_t$ is the token sampled from the vocabulary $V$.\newline
\textbf{Sampling: } For an input sentence $x$, we generate an output sentence by sampling from the vocabulary at each timestep $t$. Each output token (action) is sampled from the model's softmax distribution over the vocabulary (action space). We use a `top-$p$ sampling' (alternatively, `nucleus sampling')~\citep{holtzman2019curious} for the same. This sampling method samples a token from the set of top tokens that make up a cumulative softmax probability of $p$. For each input sentence $x$ in the RL-training set, we generate $K$ different outputs by repeating the above process $K$ times to get better estimation for the return.\newline
\textbf{Policy Gradient: }  We use the REINFORCE ~\citep{williams1992simple} policy gradient algorithm. REINFORCE relies on an estimated return by Monte-Carlo methods, using episode samples to update the policy parameter $\theta$. 
\begin{equation}
    \label{long term reward}
    \nabla_{\theta}J(\theta) = \E_\pi[\nabla_{\theta} log_e \pi_\theta(y_t|s_t;\theta) * (Q^\pi(s_t, y_t)-b)]
\end{equation}
Here $Q^\pi(s_t, y_t)$ is the action-value function and is equal to the expected return of the state-action pair $(s_t, y_t)$. In the case of REINFORCE, it is estimated from the samples collected with the current policy $\pi_\theta$. $b$ is the baseline, which helps to reduce the variance in return estimation. ~\citet{williams1992simple} discuss different methods to calculate baseline. We use a constant baseline, which is the average of $Q^{\pi}(s_t, y_t)$ over all generated tokens during the batch gradient update.
\begin{equation}
    \label{action value function}
    Q^\pi(s_t, y_t) = \E_\pi[r_{t} + \sum_{i=t+1}^{T} \gamma^{i-t} * r_{i}]
\end{equation}
Here, $T$ is the episode length, $\gamma \in [0,1]$ is the discount factor and $r_{t}$ is the reward associated with $(s_t,y_t)$. 
\subsection{Rewards}
The model is provided with token-level dense rewards that indicate: 1) how well the generated text reflects target style, 2) how much of the content it preserves from the input, and 3) fluency. The final reward for each token is the weighted sum of these individual rewards. Separate rewards have also been used by~\citet{ijcai2019-711} and~\citet{gong2019reinforcement}.

\subsubsection{Target Style Reward}
We train a BERT style-classifier with weights $\theta_{CLS}$ and use its self-attention scores to provide a reward for the target style. The classifier has multiple attention heads.~\citet{Sudhakar2019TransformingDR} isolate a single attention-head such that the self-attention scores $\bm{\alpha}$ of this head correspond to how much a token contributes to the style of the output. To do this, they use a `leave-one-out' approach on the dev set, computing $\bm{\alpha}$ for each head, and then removing the top $\floor{\lambda*|x|}$ tokens based on $\bm{\alpha}$, from the input sentence $x$. Here $\lambda \in [0,1]$ is a parameter tuned for each dataset and $|x|$ represents number of tokens in the sentence $x$. The head for which the classifier's score deviates the most is the one which~\citet{Sudhakar2019TransformingDR} used to disentangle style and content tokens. 

Using the same procedure, we arrive at a candidate head, and use its self-attention scores $\bm{\alpha}$ to assign a token level reward $rs_t$. Formally, if the attention score associated with token $y_t$ is $\alpha_{y_{t}}$ then the style reward for each token $y_t$ of the generated sentence $y$ is,
\begin{equation}
    \label{eq:style_rewards}
    rs_t = mask * (P(s^{tgt}| y; \theta_{CLS}) - 0.5)
\end{equation}
where,
\begin{equation}
    \label{style_multiplier_equation}
    mask = 
    \begin{cases}
    1, & \text{if }\ y_{t} \text{ in top $\floor{\lambda|y|}$ tokens} \\
    & \text{based on } \alpha_{y_t} \\
    0, & \text{else}
    \end{cases}
\end{equation}
The tokens which influence the style of the sentence will be rewarded with $P(s^{tgt}| y; \theta_{CLS}) - 0.5$, and purely content related tokens will receive a reward of 0. 
\subsubsection{Content Preservation Reward}
We use an n-gram based content reward to encourage the model to preserve content from the input. Let us define $G_{t}$ as the set of n-grams up to order 3 (set heuristically) associated with token $y_t$ of the generated sentence $y$ and $c(x_t)$ is the context window for the $t^{th}$ token in the input sentence $x$.
\begin{multline}
\label{Gt}
G_{t} = \{\{{y}_{t}\}, \{{y}_{t-1}, {y}_{t}\},
\{{y}_{t}, {y}_{t+1}\}, \\
\{{y}_{t-2},
{y}_{t-1},
{y}_{t}\},
\{{y}_{t},
{y}_{t+1},
{y}_{t+2}\}
\}
\end{multline}
\begin{gather}
    c(x_t) = \{x_{t-2}, x_{t-1}, x_{t}, x_{t+1}, x_{t+2}\} \\
r_{cont}(z) =
    \begin{cases}
      +1, & \text{if}\ z \text{ is present in } c(x)  \\
      -1, & \text{else}\
    \end{cases}
\end{gather}
The context preservation reward $rc_t$ for token $y_t$ is,
\begin{equation}
    \label{eq:content_rewards}
    rc_t = (1 - mask) * \frac{\sum_{g \in G_t} r_{cont}(g)}{|G_t|}
\end{equation}
Where $mask$ is as described in (\ref{style_multiplier_equation}). The term $(1 - mask)$ emphasises the fact that content preservation reward should apply to content tokens only and style tokens should not receive negative rewards for not preserving the content.

\subsubsection{Fluency Reward}
The fluency reward should apply to both attributes and content words. It should ensure that they are coherently used in context.~\citet{Li2018DeleteRG} observe that content preservation metrics (such as BLEU) do not correlate with human scores for fluency. Hence a separate fluency reward is necessary. We fine-tuned a pre-trained GPT ~\citep{radford2018improving} model on the training data with language modeling (LM) task. We then use it to determine the fluency of each token $y_t$ using (\ref{eq:fluency_rewards}). We note here that~\citet{d2019training}, use a similar dense reward to ours. We design the fluency reward as the LM likelihood:
\begin{equation}
\label{eq:fluency_rewards} 
rf_{t}  = P(y_{t}|y_{1:t-1}; \theta_{LM})
\end{equation}
\subsubsection{Overall Reward: }
\begin{figure}[h]
\includegraphics[width=7cm, height=5cm]{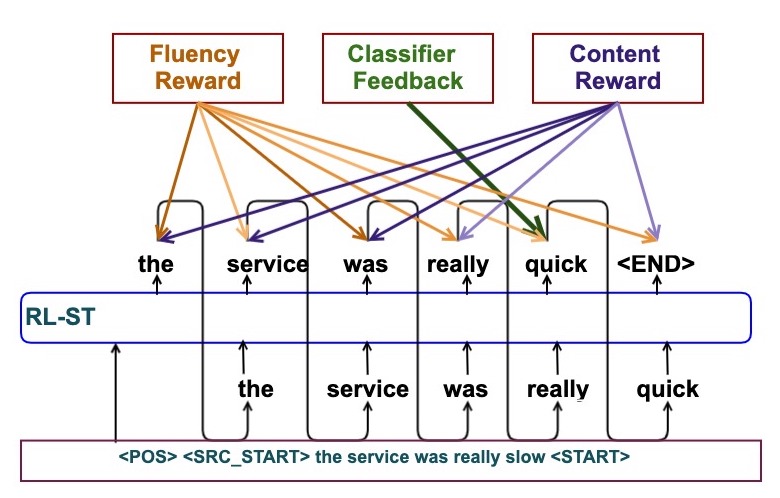}
\caption{An example from YELP for the task of sentiment transfer. \small{\textless POS\textgreater}~represents positive target style, \small{\textless SRC\_START\textgreater}~indicates the start of the input sentence, \small{\textless START\textgreater}~indicates the start of the output sentence and \small{\textless END\textgreater}~is end-of-sentence marker.}
\label{end2end}
\end{figure}

The overall reward ($r_t$) for each token is a weighted sum of the style, content and fluency rewards from equations \ref{eq:style_rewards}, \ref{eq:content_rewards} and \ref{eq:fluency_rewards}:
\begin{equation} \label{eq:overall_rewards} 
r_{t} = \lambda_{S} * rs_{t}  + \lambda_{C} * rc_{t}  +  \lambda_{F} * rf_{t}
\end{equation}
$\lambda_{S}$, $\lambda_{C}$ and $\lambda_{F}$ are reward weights. We heuristically set $\lambda_{S} = 2.0$, $\lambda_{C} = 1.0$ and $\lambda_{F} = 0.5$. Figure~\ref{end2end} shows a training example with input representation, rewards and decoding. The details of the RL training procedure are mentioned in Algorithm \ref{Algo: rl training procedure}.
\begin{algorithm}
\caption{RL training Procedure}
\label{Algo: rl training procedure}
\SetAlgoLined
 \textbf{Initialize:} pre-trained decoder model $f(\theta)$ with weights $\theta$;
reward-change threshold $\epsilon$; initial baseline $b'$; $train$ = $True$\\
 \While{train is True}{
    Sample N sentences from training set $D$\;
    \For{each sentence}{
        Generate K output sentences\;
        \For{each token $y_t$ of output sentence}{
            calculate style reward $rs_t$ \ref{eq:style_rewards}\;
            calculate content reward $rc_t$ \ref{eq:content_rewards}\;
            calculate fluency reward $rf_t$ \ref{eq:fluency_rewards}\;
            calculate total reward $rt$ \ref{eq:overall_rewards}\;
            calculate return $Q^{\pi}(s_t, y_t)$ \ref{action value function}\;
        }
    }
    previous_baseline $b' = b$\;
    baseline $b = \E[Q^{\pi}(s,y)]$\;
    calculate $\nabla_{\theta}J(\theta)$ from \ref{long term reward}\;
    Update $\theta$ based on $\nabla_{\theta}J(\theta)$;
    
    \uIf{$|b - b'| < \epsilon$}{
    $train$ = $False$\;
  }
 }
\end{algorithm}
\subsection{Return Estimation}
It must be mentioned here that in previous work~\citep{ijcai2019-711, gong2019reinforcement}, the value of the discount factor $\gamma$ is set to a non-zero value in eq. \ref{action value function}. This is to allow for estimation of token-level return from sparse rewards. However, the proposed method of designing dense rewards, solves the credit assignment problem directly. Consequently, there is no dependency on future rewards to estimate $Q^{\pi}(s_t, y_t)$. Hence, we set the value of $\gamma$ to 0.
\section{Experiments}
\subsection{Datasets} \label{sec:dataset_stats}
Following are brief descriptions of the datasets we use, borrowed from works that use them previously.\newline
\textbf{YELP: } Each sentence is from a business review on Yelp, and is labeled as having either positive or negative sentiment~\citep{Li2018DeleteRG}. The task is to transfer positive sentences to negative sentences and vice-versa.~\citet{Li2018DeleteRG} publish a set of human reference outputs on the test set of YELP.\newline
\textbf{GYAFC: } Each sentence is labeled as either being formal or informal. We only use a subset of this dataset which corresponds to Yahoo answers in the Family and Relationships domain. The task is to transfer formal to informal sentences and vice-versa.~\citet{rao-tetreault-2018-dear} publish a set of human reference outputs on the test set of GYAFC. Though GYAFC has source and target sentences aligned, we discard these alignments, since our problem setting is unsupervised.\newline
Table \ref{dataset-stats} shows train, dev and test statistics of the datasets.
\begin{table}[h]
\begin{center}
\small
\begin{tabular}{|c|c|c|c|c|}
\hline \bf Dataset & \bf Style & \bf Train & \bf Dev & \bf Test \\ \hline
YELP & Positive & 270K & 2000 & 500 \\
& Negative & 180K & 2000 & 500 \\
\hline
GYAFC & Formal & 277K & 985 & 1019 \\
& Informal & 279K & 1015 & 1332 \\
\hline
\end{tabular}
\end{center}
\caption{\label{dataset-stats} Dataset statistics}
\end{table}
\subsection{Baseline Models} \label{sec:baseline models}
A baseline model refers to a pre-trained version of DRL, as described in section~\ref{sec:mle}. We pre-train three different baseline models - DO\textsubscript{B}, BGST-S\textsubscript{B} and BGST\textsubscript{B}. Baseline models of varying quality are deliberately chosen, so that we can observe the contribution of RL training over each of them independently.
The parallel corpus for a) DO\textsubscript{B} is obtained using the DeleteOnly model (DO)~\citep{Li2018DeleteRG}, b) for BGST\textsubscript{B}, using the Blind Generative Style Transformer (BGST)~\citep{Sudhakar2019TransformingDR}, and c) for BGST-S\textsubscript{B}, using BGST-Small, which is BGST trained over a subset of the training data (5\%).\newline

\subsection{Reproducibility}
All models described below are used from Huggingface\footnote{https://github.com/huggingface/transformers}~\cite{wolf-etal-2020-transformers}.\newline
\textbf{Style Classifier:} We train the BERT-base uncased model with the default hyper-parameter setting.\newline
\textbf{Generator:}
We use GPT~\citep{radford2018improving} as the generator model. The Generator training is done in two phases, 1) LM pre-training and 2) RL training.\newline
\textbf{Pre-training:} We train over GPT's pre-trained model for our pseudo-supervised pre-training\newline
\textbf{RL-training:} We set learning rate to $lr=6.25e-6$, and gradient clipping to $1$ to avoid overflow. Other hyper-parameters are set to the default values provided.
\subsection{Evaluation Methods}
\subsubsection{Human Evaluation}
We obtain human evaluations of model outputs from crowd workers on MTurk. For each source
sentence and target style, we present the outputs of all models being compared, to the same workers. They are all native English speakers from North America and were made familiar with the datasets. They were asked to rate each output on the following three criteria, each on a Likert scale from 1 to 5: a) target style match (Sty.), b) content preservation (Con.), c) fluency (Flu.). We also calculate the overall success rate (All) as the percentage of times a model received either a 4 or a 5 on all the above three criteria.
\subsubsection{Automatic Evaluation}
To measure target style strength (Sty.) of outputs, we use a pre-trained FastText\footnote{https://fasttext.cc/}~\citep{joulin2017bag} style classifiers for each dataset, which achieve accuracies of 96.5\% and 89.5\% on the test sets of YELP and GYAFC respectively. For content preservation (Con.), we calculate the average BLEU scores of the output with respect to the human reference sentences. Fluency (Flu.) is estimated by finetuning pre-trained OpenAI GPT-2~\citep{radford2019language} models (different from any of the GPT models used in this work) on each dataset's training set, and using them to obtain the perplexity of the output sentences. They achieve perplexities of 21.42 and 52.5 on the test sets of YELP and GYAFC respectively. We also calculate the geometric mean, GM (All), of style and content scores. It must be noted, however, that most previous works such as~\citet{Li2018DeleteRG} and~\citet{Sudhakar2019TransformingDR} note that human evaluations are more reliable than automatic evaluations. Hence, in all further analysis in this work, we draw conclusions from the human evaluations.

\section{Results}
\subsection{Improvement over Baseline Models}
As mentioned in section~\ref{sec:baseline models}, we further train each baseline model using dense RL rewards. We observe that our novel RL training provides significant improvement over each of these baseline models. To further distill the effects of RL separately, we consider an additional simple baseline model (SOURCE\textsubscript{B}), trained on a parallel corpus where the target sentence is the same as the source sentence. Human evaluations of these models are presented in Table~\ref{human-eval-baselines}. These evaluations show that DRL significantly improves over a variety of baseline (BL) models on style, content as well as fluency, irrespective of their original style transfer capabilities.

\subsection{Comparison with Roll-out}
\label{roll-out-comparison}
We compare dense rewards with the roll-out strategy. To do so, we train a RL style transfer model which uses the roll-out strategy for reward shaping, as used by~\citet{gong2019reinforcement}. We keep all other factors such as pre-training and model architecture same.
We compute the rolled out reward $r_t$  using the same process as~\citet{gong2019reinforcement}. We set $mask = 1$ in eq. \ref{eq:style_rewards} to calculate style reward $rs_t$ and $mask = 0$ in eq. \ref{eq:content_rewards} to calculate content reward $rc_t$. We set $\gamma = 1$ while calculating estimated return $Q^{\pi}(s_t, y_t)$ using equation \ref{action value function}.\newline 
This allows us to train a Roll-Out model (RO) which a) has the same architecture as DRL described in section~\ref{approach} and b) is pre-trained with a synthetic corpus generated by the BGST-S\textsubscript{B} baseline.\newline\newline
\textbf{Analysis: }
We compare a) RO, b) its baseline model BGST-S\textsubscript{B} and c) DRL trained over the same baseline model. These results are presented in Table~\ref{human-eval-baselines}. The results indicate the superior performance of dense rewards in comparison to the roll-out strategy.
\subsubsection{Sample Efficiency}
\begin{figure}[h]
\includegraphics[width=7cm, height=6cm]{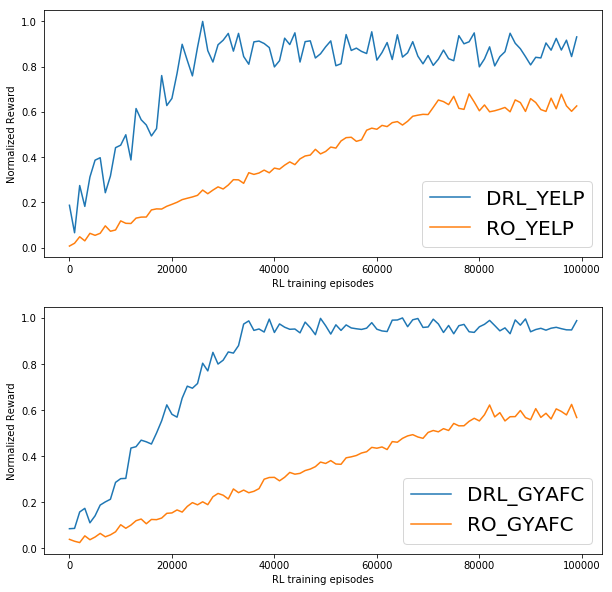}
\caption{Training steps}
\label{sample efficiency}
\end{figure}

In Figure \ref{sample efficiency}, we compare the sample efficiencies of our DRL and RO. The overall reward (\ref{eq:overall_rewards}) is normalized between 0 to 1 for easy visualization as the training steps progress. 
The peak reward of DRL is higher than RO in both cases. On YELP, DRL reaches its peak reward after 25K episodes whereas RO takes 74K episodes. On GYAFC, DRL takes 40K episodes, whereas RO takes 83K. Since all other factors are kept same in both setups, these gains can be attributed to accurate dense rewards for each token, resulting in better estimation of $Q^\pi(s_t, y_t)$, sample efficiency and convergence of the policy.\newline
\subsubsection{Training Time}
The roll-out itself is an expensive operation, requiring $O(T^{2})$ computations for one complete output, where $T$ is the output length. Time complexity for self-attention based reward shaping is $O(T)$. In practice, training DRL is 4x faster than RO on YELP, and 10x faster on GYAFC, owing to the differences in average sentence length.
\subsection{Comparison with Fully Attention-based Reward}
\label{continuous-section}
As an additional experiment, we explore directly using attention scores in the style and content rewards, without explicitly separating attributes and content. We modify the reward coefficient $mask$ in equation~\ref{style_multiplier_equation} (which will reflect in the style and content rewards) to the attention weight as: 
$mask = \alpha_{t}$ and train the attention-reward model (DRL-At) over the BGST-S\textsubscript{B} baseline model, keeping all else the same as DRL.\newline\newline
\textbf{Analysis: }We compare a) DRL-At, b) its baseline model BGST-S\textsubscript{B} and c) DRL trained over the same baseline model. These results are presented in Table~\ref{human-eval-baselines}. While DRL-At does give encouraging results, we see that it still performs narrowly worse than DRL. We surmise that this might be because a) every word is assigned both a content and style reward, even though there exists a hard separation between content and attributes in most cases, and b) raw attention scores might not be the most appropriate reward coefficients.
\subsection{Comparison with Previous Works}\label{comparison-prev}
We compare our results with previous works that use different approaches for style transfer. We choose the state-of-art work from each approach. Multi Decoder (\textbf{MD})~\citep{fu2018style} uses adversarial training to separate content and attribute. Back-Translation (\textbf{BT})~\citep{style_transfer_acl18} uses back-translation that has reduced style, and then generate style-specific outputs. We also compare with RL based approaches described in previous sections. UnPaired (\textbf{UnP})~\citep{xu-etal-2018-unpaired} uses a cyclic RL approach, RL-roll-out (\textbf{RL-RO})~\citep{gong2019reinforcement} uses the roll-out strategy and DualRL (\textbf{DuR})~\citep{ijcai2019-711} uses a naive reward shaping. Finally, \textbf{DO}~\citep{Li2018DeleteRG} and \textbf{BGST}~\citep{Sudhakar2019TransformingDR} both separate content and attributes, and train a generative model to reconstruct the source sentence, given only the content. Evaluations of these works and ours are presented in Table~\ref{human-eval}. In this table, DRL refers to DRL trained over the BGST\textsubscript{B} baseline model, since it is our best performing from Table~\ref{human-eval-baselines}.\newline\newline
\textbf{Analysis: }
We see that DRL improves over previous state-of-art models by a good margin on all metrics across all datasets, according to both human and automatic evaluations. On the overall scores (All), we improve over previous state-of-the art model (DuR) by 21\% each on YELP and GYAFC, according to human evaluation in Table \ref{human-eval}. From manual inspection we observe that DRL performs better than previous state-of-art models in the following ways: 1) maintains consistent context across longer sentences, 2) maintains consistency of style even for sentences having multiple attribute words, 3) produces output sentences having appropriate and meaningful attributes, many of which the model has not observed at training time, and 4) does away with redundancy and repetition of words in output sentences, which is commonly observed in previous works.\newline\newline
\textbf{Sentiment versus formality:} The last row in Table~\ref{human-eval} shows the performance of humans (H) for each of the datasets. Humans perform better on YELP as compared to GYAFC, according to human evaluations. Further, through manual annotation from crowd workers on Mturk\footnote{https://www.mturk.com/}, we observe that the average percentage of word changes to generate target style sentence is 23\% for YELP, and 39\% for GYAFC. The average sentence length is 8 words in YELP and 13 words in GYAFC. These statistics, along with manual investigations suggest that formality transfer on GYAFC is a more complex and ambiguous task than sentiment transfer on YELP. The efficacy of using dense rewards is further shown by the fact that DRL is only 6\% worse than humans on GYAFC.\newline\newline
Table ~\ref{sents-examples} shows example outputs generated by our models from the test set.

\begin{table*}[h]
\begin{center}
\small
\begin{tabular}{|c|c|cccc|cccc|}
\hline
\multicolumn{2}{|c|}{} & \multicolumn{4}{|c|}{\textbf{YELP}} & \multicolumn{4}{|c|}{\textbf{GYAFC}} \\
\hline
\hline
\textbf{Baseline (BL)} & \textbf{Model} & \textbf{Sty.} & \textbf{Con.} & \textbf{Flu.} & \textbf{All} & \textbf{Sty.} & \textbf{Con.} & \textbf{Flu.} & \textbf{All} \\
\hline
\hline
SOURCE\textsubscript{B} & BL & 0.71 & 4.61 & 4.67 & 2\% & 0.59 & 4.12 & 3.83 & 1\%  \\
& \textbf{DRL} & \textbf{3.27} & \textbf{3.46} & \textbf{3.52} & \textbf{14\%} & \textbf{3.05} & \textbf{3.11} & \textbf{3.09} & \textbf{11\%} \\
\hline
DO\textsubscript{B} & BL & 3.18 & 3.27 & 3.21 & 19\% & 2.97 & 3.05 & 2.86 & 8\% \\
& \textbf{DRL} & \textbf{4.27} & \textbf{4.12} & \textbf{4.31} & \textbf{37\%} & \textbf{3.99} & \textbf{4.03} & \textbf{3.92} & \textbf{22\%} \\
\hline
BGST-S\textsubscript{B} & BL & 3.30 & 3.50 & 3.67 & 24\% & 3.07 & 3.16 & 3.24 & 13\% \\
& RO & 3.61 & 3.64 & 3.68 & 29\% & 3.40 & 3.28 & 3.37 & 17\% \\
& DRL-At & 4.21 & 4.28 & 4.25 & 42\% & 4.18 & 4.25 & 4.20 & 37\% \\
& \textbf{DRL} & \textbf{4.43} & \textbf{4.46} & \textbf{4.46} & \textbf{48\%} & \textbf{4.28} & \textbf{4.32} & \textbf{4.31} & \textbf{40\%} \\
\hline
BGST\textsubscript{B} & BL & 4.03 & 3.98 & 3.87 & 36\% & 3.21 & 3.24 & 3.17 & 18\% \\
& \textbf{DRL} & \textbf{4.60} & \textbf{4.68} & \textbf{4.64} & \textbf{60\%} & \textbf{4.33} & \textbf{4.45} & \textbf{4.40} & \textbf{47\%} \\
\hline
\end{tabular}
\end{center}
\caption{Human evaluation results of improvements over baselines, roll-outs, and fully attention-based rewards: Sty. = Target Style Match; Con. = Content Preservation; Flu. = Fluency; All = Overall success rate. DRL in each cell denotes DRL trained over the corresponding baseline.}
\label{human-eval-baselines}
\end{table*}

\begin{table*}[h]
\begin{center}
\scriptsize
\begin{tabular}{|c||cccc|cccc||cccc|cccc|}
\hline
& \multicolumn{8}{|c||}{\textbf{HUMAN EVALUATION}} & \multicolumn{8}{|c|}{\textbf{AUTOMATIC EVALUATION}} \\
\hline
& \multicolumn{4}{|c|}{\textbf{YELP}} & \multicolumn{4}{|c||}{\textbf{GYAFC}} & \multicolumn{4}{|c|}{\textbf{YELP}} & \multicolumn{4}{|c|}{\textbf{GYAFC}} \\
\hline
\hline
\textbf{Model} & \textbf{Sty.} & \textbf{Con.} & \textbf{Flu.} & \textbf{All} & \textbf{Sty.} & \textbf{Con.} & \textbf{Flu.} & \textbf{All} & \textbf{Sty.} & \textbf{Con.} & \textbf{Flu.$\downarrow$} & \textbf{All} & \textbf{Sty.} & \textbf{Con.} & \textbf{Flu.$\downarrow$} & \textbf{All} \\
\hline
\hline
MD & 2.07 & 3.11 & 3.19 & 6\% & 2.17 & 1.89 & 2.29 & 5\% & 50.6 & 42.9 & 205.5 & 46.6 & 26.1 & 26.8 & 231.5 & 26.4 \\
BT & 4.13 & 2.41 & 4.09 & 7\% & 3.04 & 1.91 & 3.21 & 9\% & 95.5 & 25.6 & \textbf{28.5} & 49.4 & 51.5 & 18.7 & 147.2 & 31.0 \\
\hline
DO & 3.36 & 3.51 & 3.68 & 25\% & 2.31 & 2.47 & 2.84 & 7\% & 85.9 & 44.4 & 75.8 & 61.7 & 26.0 & 42.4 & 183.1 & 33.2\\
BGST & 3.78 & 3.89 & 3.82 & 36\% & 3.01 & 3.16 & 3.11 & 14\% & 86.7 & 57.1 & 38.6 & 70.4 & 69.0 & 47.4 & 89.0 & 57.2\\
\hline
UnP & 2.86 & 3.52 & 3.28 & 14\% & 2.74 & 1.45 & 2.63 & 4\% & 53.3 & 46.3 & 294.2 & 49.7 & 68.5 & 12.3 & 133.9 & 29.0 \\
RL-RO & 2.93 & 3.41 & 3.72 & 15\% & 2.83 & 1.75 & 2.61 & 6\% & 55.7 & 47.2 & 177 & 50.2 & 69.2 & 14.8 & 141.9 & 31.2\\
DuR & 3.92 & 3.89 & 4.01 & 42\% & 3.28 & 3.62 & 3.87 & 27\% & 89.2 & 59.5 & 53.2 & 72.9 & 60.1 & 44.3 & 321.1 & 51.6\\
\hline
\textbf{DRL} & \textbf{4.45} & \textbf{4.41} & \textbf{4.24} & \textbf{63\%} & \textbf{4.32} & \textbf{3.58} & \textbf{4.10} & \textbf{48\%} & \textbf{96.8} & \textbf{59.6} & 40.5 & \textbf{75.7} & \textbf{98.9} & \textbf{47.7} & \textbf{27.9} & \textbf{68.5} \\
\hline
\hline
H & 4.79 & 4.67 & 4.61 & 81\% & 4.65 & 4.38 & 4.42 & 54\% & 75.0 & 70.3 & 57.7 & 72.5 & 86.4 & 65.4 & 91.9 & 75.2 \\
\hline
\end{tabular}
\end{center}
\caption{Human and automatic evaluation results of comparison with previous works: Sty. = Target Style Match; Con. = Content Preservation; Flu. = Fluency; All (human evaluation) = Overall success rate ; All (automatic evaluation) = GM of Sty. and Con. of automatic evaluation. Fluency for automatic metrics is measured by perplexity, so the lower the better. DRL refers to DRL trained over the BGST\textsubscript{B} baseline model.}
\label{human-eval}
\end{table*}
\begin{table}[h]
\begin{center}
\small
\begin{tabular}{|c|}
\hline
\textbf{YELP ({\color{blue} Positive} to {\color{red} Negative})} \\
\hline
steve was {\color{blue}professional} and {\color{blue}found exactly the right unit } \\ to fit in our space \\
\textbf{steve was {\color{red} rude} and {\color{red}didn't have the right unit }} \\ \textbf{to fit in our space .} \\
\hline
\hline
\textbf{YELP ({\color{red} Negative} to {\color{blue} Positive})}\\
\hline
they {\color{red} tried to take advantage} of me because i am young . \\
\textbf{they {\color{blue}take great care of} me because i am young .} \\
\hline
\hline
\textbf{GYAFC ({\color{blue} Formal} to {\color{red} Informal})} \\
\hline
{\color{blue}do not} approach her and let her know that you \\ {\color{blue}find her looks very attractive.} \\
\textbf{{\color{red}don't} approach her and let her know that you} \\ \textbf{{\color{red}like her} .}\\
\hline
\hline
\textbf{GYAFC ({\color{red} Informal} to {\color{blue} Formal})} \\
\hline
{\color{red}well} that is just the way it is {\color{red}I guess. }\\
{\textbf{that is just the way it is , {\color{blue}i would advise} .}}  \\
\hline
\end{tabular}
\end{center}
\caption{Examples of generated sentences by DRL. Each cell has the source sentence first and the generated sentence second.}
\label{sents-examples}
\end{table}
\section{Conclusion}
We present a novel method to provide dense rewards for unsupervised text generation and compare it with the currently used roll-out strategy using ablation studies. We demonstrate the use of these dense rewards in building a state-of-art text style transfer system.

\bibliography{anthology,custom}

\begin{thebibliography}{36}
\expandafter\ifx\csname natexlab\endcsname\relax\def\natexlab#1{#1}\fi

\bibitem[{Bahdanau et~al.(2017)Bahdanau, Brakel, Xu, Goyal, Lowe, Pineau,
  Courville, and Bengio}]{Bahdanau2017AnAA}
Dzmitry Bahdanau, Philemon Brakel, Kelvin Xu, Anirudh Goyal, Ryan Lowe, Joelle
  Pineau, Aaron~C. Courville, and Yoshua Bengio. 2017.
\newblock An actor-critic algorithm for sequence prediction.
\newblock \emph{ArXiv}, abs/1607.07086.

\bibitem[{d'Autume et~al.(2019)d'Autume, Rosca, Rae, and
  Mohamed}]{d2019training}
Cyprien de~Masson d'Autume, Mihaela Rosca, Jack Rae, and Shakir Mohamed. 2019.
\newblock Training language gans from scratch.
\newblock \emph{arXiv preprint arXiv:1905.09922}.

\bibitem[{Devlin et~al.(2018)Devlin, Chang, Lee, and
  Toutanova}]{devlin2018bert}
Jacob Devlin, Ming-Wei Chang, Kenton Lee, and Kristina Toutanova. 2018.
\newblock Bert: Pre-training of deep bidirectional transformers for language
  understanding.
\newblock \emph{arXiv preprint arXiv:1810.04805}.

\bibitem[{Fu et~al.(2018)Fu, Tan, Peng, Zhao, and Yan}]{fu2018style}
Zhenxin Fu, Xiaoye Tan, Nanyun Peng, Dongyan Zhao, and Rui Yan. 2018.
\newblock Style transfer in text: Exploration and evaluation.
\newblock In \emph{Thirty-Second AAAI Conference on Artificial Intelligence}.

\bibitem[{Gong et~al.(2019)Gong, Bhat, Wu, Xiong, and mei
  Hwu}]{gong2019reinforcement}
Hongyu Gong, Suma Bhat, Lingfei Wu, Jinjun Xiong, and Wen mei Hwu. 2019.
\newblock \href {http://arxiv.org/abs/1903.10671} {Reinforcement learning based
  text style transfer without parallel training corpus}.

\bibitem[{Holtzman et~al.(2019)Holtzman, Buys, Forbes, and
  Choi}]{holtzman2019curious}
Ari Holtzman, Jan Buys, Maxwell Forbes, and Yejin Choi. 2019.
\newblock \href {http://arxiv.org/abs/1904.09751} {The curious case of neural
  text degeneration}.

\bibitem[{Hu et~al.(2017)Hu, Yang, Liang, Salakhutdinov, and
  Xing}]{pmlr-v70-hu17e}
Zhiting Hu, Zichao Yang, Xiaodan Liang, Ruslan Salakhutdinov, and Eric~P. Xing.
  2017.
\newblock \href {http://proceedings.mlr.press/v70/hu17e.html} {Toward
  controlled generation of text}.
\newblock In \emph{Proceedings of the 34th International Conference on Machine
  Learning}, volume~70 of \emph{Proceedings of Machine Learning Research},
  pages 1587--1596, International Convention Centre, Sydney, Australia. PMLR.

\bibitem[{John et~al.(2019)John, Mou, Bahuleyan, and
  Vechtomova}]{john-etal-2019-disentangled}
Vineet John, Lili Mou, Hareesh Bahuleyan, and Olga Vechtomova. 2019.
\newblock \href {https://doi.org/10.18653/v1/P19-1041} {Disentangled
  representation learning for non-parallel text style transfer}.
\newblock In \emph{Proceedings of the 57th Annual Meeting of the Association
  for Computational Linguistics}, pages 424--434, Florence, Italy. Association
  for Computational Linguistics.

\bibitem[{Joulin et~al.(2017)Joulin, Grave, Bojanowski, and
  Mikolov}]{joulin2017bag}
Armand Joulin, Edouard Grave, Piotr Bojanowski, and Tomas Mikolov. 2017.
\newblock Bag of tricks for efficient text classification.
\newblock In \emph{Proceedings of the 15th Conference of the European Chapter
  of the Association for Computational Linguistics: Volume 2, Short Papers},
  pages 427--431. Association for Computational Linguistics.

\bibitem[{Kocsis and Szepesv{\'a}ri(2006)}]{mcts}
Levente Kocsis and Csaba Szepesv{\'a}ri. 2006.
\newblock Bandit based monte-carlo planning.
\newblock In \emph{Machine Learning: ECML 2006}, pages 282--293, Berlin,
  Heidelberg. Springer Berlin Heidelberg.

\bibitem[{Lample et~al.(2018)Lample, Ott, Conneau, Denoyer, and
  Ranzato}]{lample-etal-2018-phrase}
Guillaume Lample, Myle Ott, Alexis Conneau, Ludovic Denoyer, and Marc{'}Aurelio
  Ranzato. 2018.
\newblock \href {https://www.aclweb.org/anthology/D18-1549} {Phrase-based {\&}
  neural unsupervised machine translation}.
\newblock In \emph{Proceedings of the 2018 Conference on Empirical Methods in
  Natural Language Processing}, pages 5039--5049, Brussels, Belgium.
  Association for Computational Linguistics.

\bibitem[{Li et~al.(2018{\natexlab{a}})Li, Jia, He, and
  Liang}]{li-etal-2018-delete}
Juncen Li, Robin Jia, He~He, and Percy Liang. 2018{\natexlab{a}}.
\newblock \href {https://doi.org/10.18653/v1/N18-1169} {Delete, retrieve,
  generate: a simple approach to sentiment and style transfer}.
\newblock In \emph{Proceedings of the 2018 Conference of the North {A}merican
  Chapter of the Association for Computational Linguistics: Human Language
  Technologies, Volume 1 (Long Papers)}, pages 1865--1874, New Orleans,
  Louisiana. Association for Computational Linguistics.

\bibitem[{Li et~al.(2018{\natexlab{b}})Li, Jia, He, and Liang}]{Li2018DeleteRG}
Juncen Li, Robin Jia, He~He, and Percy Liang. 2018{\natexlab{b}}.
\newblock Delete, retrieve, generate: A simple approach to sentiment and style
  transfer.
\newblock \emph{ArXiv}, abs/1804.06437.

\bibitem[{Li et~al.(2018{\natexlab{c}})Li, Jiang, Shang, and
  Li}]{li-etal-2018-paraphrase}
Zichao Li, Xin Jiang, Lifeng Shang, and Hang Li. 2018{\natexlab{c}}.
\newblock \href {https://doi.org/10.18653/v1/D18-1421} {Paraphrase generation
  with deep reinforcement learning}.
\newblock In \emph{Proceedings of the 2018 Conference on Empirical Methods in
  Natural Language Processing}, pages 3865--3878, Brussels, Belgium.
  Association for Computational Linguistics.

\bibitem[{Liu et~al.(2019)Liu, Fu, Zhang, Pal, and Lv}]{liu2019revision}
Dayiheng Liu, Jie Fu, Yidan Zhang, Chris Pal, and Jiancheng Lv. 2019.
\newblock Revision in continuous space: Fine-grained control of text style
  transfer.
\newblock \emph{arXiv preprint arXiv:1905.12304}.

\bibitem[{Luo et~al.(2019)Luo, Li, Zhou, Yang, Chang, Sun, and
  Sui}]{ijcai2019-711}
Fuli Luo, Peng Li, Jie Zhou, Pengcheng Yang, Baobao Chang, Xu~Sun, and Zhifang
  Sui. 2019.
\newblock \href {https://doi.org/10.24963/ijcai.2019/711} {A dual reinforcement
  learning framework for unsupervised text style transfer}.
\newblock In \emph{Proceedings of the Twenty-Eighth International Joint
  Conference on Artificial Intelligence, {IJCAI-19}}, pages 5116--5122.
  International Joint Conferences on Artificial Intelligence Organization.

\bibitem[{Papineni et~al.(2002)Papineni, Roukos, Ward, and
  Zhu}]{papineni-etal-2002-bleu}
Kishore Papineni, Salim Roukos, Todd Ward, and Wei-Jing Zhu. 2002.
\newblock \href {https://doi.org/10.3115/1073083.1073135} {{B}leu: a method for
  automatic evaluation of machine translation}.
\newblock In \emph{Proceedings of the 40th Annual Meeting of the Association
  for Computational Linguistics}, pages 311--318, Philadelphia, Pennsylvania,
  USA. Association for Computational Linguistics.

\bibitem[{Paulus et~al.(2018)Paulus, Xiong, and Socher}]{Paulus2018ADR}
Romain Paulus, Caiming Xiong, and R.~Socher. 2018.
\newblock A deep reinforced model for abstractive summarization.
\newblock \emph{ArXiv}, abs/1705.04304.

\bibitem[{Prabhumoye et~al.(2018)Prabhumoye, Tsvetkov, Salakhutdinov, and
  Black}]{style_transfer_acl18}
Shrimai Prabhumoye, Yulia Tsvetkov, Ruslan Salakhutdinov, and Alan~W Black.
  2018.
\newblock Style transfer through back-translation.
\newblock In \emph{Proc. ACL}.

\bibitem[{Radford et~al.()Radford, Narasimhan, Salimans, and
  Sutskever}]{radford2018improving}
Alec Radford, Karthik Narasimhan, Tim Salimans, and Ilya Sutskever.
\newblock Improving language understanding by generative pre-training.

\bibitem[{Radford et~al.(2019)Radford, Wu, Child, Luan, Amodei, and
  Sutskever}]{radford2019language}
Alec Radford, Jeff Wu, Rewon Child, David Luan, Dario Amodei, and Ilya
  Sutskever. 2019.
\newblock Language models are unsupervised multitask learners.

\bibitem[{Rao and Tetreault(2018)}]{rao-tetreault-2018-dear}
Sudha Rao and Joel Tetreault. 2018.
\newblock \href {https://doi.org/10.18653/v1/N18-1012} {Dear sir or madam, may
  {I} introduce the {GYAFC} dataset: Corpus, benchmarks and metrics for
  formality style transfer}.
\newblock In \emph{Proceedings of the 2018 Conference of the North {A}merican
  Chapter of the Association for Computational Linguistics: Human Language
  Technologies, Volume 1 (Long Papers)}, pages 129--140, New Orleans,
  Louisiana. Association for Computational Linguistics.

\bibitem[{Schmidt and Hofmann(2020)}]{Schmidt2020BERTAA}
Florian Schmidt and T.~Hofmann. 2020.
\newblock Bert as a teacher: Contextual embeddings for sequence-level reward.
\newblock \emph{ArXiv}, abs/2003.02738.

\bibitem[{Shi et~al.(2018)Shi, Chen, Qiu, and Huang}]{Shi2018TowardDT}
Zhan Shi, Xinchi Chen, Xipeng Qiu, and X.~Huang. 2018.
\newblock Toward diverse text generation with inverse reinforcement learning.
\newblock In \emph{IJCAI}.

\bibitem[{Sudhakar et~al.(2019)Sudhakar, Upadhyay, and
  Maheswaran}]{Sudhakar2019TransformingDR}
A.~Sudhakar, Bhargav Upadhyay, and A.~Maheswaran. 2019.
\newblock Transforming delete, retrieve, generate approach for controlled text
  style transfer.
\newblock \emph{ArXiv}, abs/1908.09368.

\bibitem[{Williams(1992)}]{williams1992simple}
Ronald~J Williams. 1992.
\newblock Simple statistical gradient-following algorithms for connectionist
  reinforcement learning.
\newblock \emph{Machine learning}, 8(3-4):229--256.

\bibitem[{Wolf et~al.(2020)Wolf, Debut, Sanh, Chaumond, Delangue, Moi, Cistac,
  Rault, Louf, Funtowicz, Davison, Shleifer, von Platen, Ma, Jernite, Plu, Xu,
  Scao, Gugger, Drame, Lhoest, and Rush}]{wolf-etal-2020-transformers}
Thomas Wolf, Lysandre Debut, Victor Sanh, Julien Chaumond, Clement Delangue,
  Anthony Moi, Pierric Cistac, Tim Rault, Rémi Louf, Morgan Funtowicz, Joe
  Davison, Sam Shleifer, Patrick von Platen, Clara Ma, Yacine Jernite, Julien
  Plu, Canwen Xu, Teven~Le Scao, Sylvain Gugger, Mariama Drame, Quentin Lhoest,
  and Alexander~M. Rush. 2020.
\newblock \href {https://www.aclweb.org/anthology/2020.emnlp-demos.6}
  {Transformers: State-of-the-art natural language processing}.
\newblock In \emph{Proceedings of the 2020 Conference on Empirical Methods in
  Natural Language Processing: System Demonstrations}, pages 38--45, Online.
  Association for Computational Linguistics.

\bibitem[{Wu et~al.(2019{\natexlab{a}})Wu, Ren, Luo, and
  Sun}]{wu-etal-2019-hierarchical-reinforced}
Chen Wu, Xuancheng Ren, Fuli Luo, and Xu~Sun. 2019{\natexlab{a}}.
\newblock \href {https://doi.org/10.18653/v1/P19-1482} {A hierarchical
  reinforced sequence operation method for unsupervised text style transfer}.
\newblock In \emph{Proceedings of the 57th Annual Meeting of the Association
  for Computational Linguistics}, pages 4873--4883, Florence, Italy.
  Association for Computational Linguistics.

\bibitem[{Wu et~al.(2018)Wu, Tian, Qin, Lai, and Liu}]{Wu2018ASO}
Lijun Wu, Fei Tian, Tao Qin, J.~Lai, and T.~Liu. 2018.
\newblock A study of reinforcement learning for neural machine translation.
\newblock In \emph{EMNLP}.

\bibitem[{Wu et~al.(2019{\natexlab{b}})Wu, Zhang, Zang, Han, and
  Hu}]{wu2019mask}
Xing Wu, Tao Zhang, Liangjun Zang, Jizhong Han, and Songlin Hu.
  2019{\natexlab{b}}.
\newblock " mask and infill": Applying masked language model to sentiment
  transfer.
\newblock \emph{arXiv preprint arXiv:1908.08039}.

\bibitem[{Xu et~al.(2018)Xu, SUN, Zeng, Zhang, Ren, Wang, and
  Li}]{xu-etal-2018-unpaired}
Jingjing Xu, Xu~SUN, Qi~Zeng, Xiaodong Zhang, Xuancheng Ren, Houfeng Wang, and
  Wenjie Li. 2018.
\newblock \href {https://www.aclweb.org/anthology/P18-1090} {Unpaired
  sentiment-to-sentiment translation: A cycled reinforcement learning
  approach}.
\newblock In \emph{Proceedings of the 56th Annual Meeting of the Association
  for Computational Linguistics (Volume 1: Long Papers)}, pages 979--988,
  Melbourne, Australia. Association for Computational Linguistics.

\bibitem[{Yang et~al.(2018)Yang, Hu, Dyer, Xing, and
  Berg-Kirkpatrick}]{yang2018unsupervised}
Zichao Yang, Zhiting Hu, Chris Dyer, Eric~P Xing, and Taylor Berg-Kirkpatrick.
  2018.
\newblock Unsupervised text style transfer using language models as
  discriminators.
\newblock In \emph{Advances in Neural Information Processing Systems}, pages
  7287--7298.

\bibitem[{Yu et~al.(2017)Yu, Zhang, Wang, and Yu}]{yu2017seqgan}
Lantao Yu, Weinan Zhang, Jun Wang, and Yong Yu. 2017.
\newblock Seqgan: Sequence generative adversarial nets with policy gradient.
\newblock In \emph{Thirty-First AAAI Conference on Artificial Intelligence}.

\bibitem[{Zhang and Lapata(2017)}]{zhang-lapata-2017-sentence}
Xingxing Zhang and Mirella Lapata. 2017.
\newblock \href {https://doi.org/10.18653/v1/D17-1062} {Sentence simplification
  with deep reinforcement learning}.
\newblock In \emph{Proceedings of the 2017 Conference on Empirical Methods in
  Natural Language Processing}, pages 584--594, Copenhagen, Denmark.
  Association for Computational Linguistics.

\bibitem[{Zhao et~al.(2018)Zhao, Bi, Cai, Liu, Tu, and Shi}]{zhao2018language}
Yanpeng Zhao, Wei Bi, Deng Cai, Xiaojiang Liu, Kewei Tu, and Shuming Shi. 2018.
\newblock \href {http://arxiv.org/abs/1808.04071} {Language style transfer from
  sentences with arbitrary unknown styles}.

\bibitem[{Ziebart et~al.(2008)Ziebart, Maas, Bagnell, and Dey}]{ZiebartMBD08}
Brian~D. Ziebart, Andrew~L. Maas, J.~Andrew Bagnell, and Anind~K. Dey. 2008.
\newblock \href {http://www.aaai.org/Library/AAAI/2008/aaai08-227.php} {Maximum
  entropy inverse reinforcement learning}.
\newblock In \emph{Proceedings of the Twenty-Third {AAAI} Conference on
  Artificial Intelligence, {AAAI} 2008, Chicago, Illinois, USA, July 13-17,
  2008}, pages 1433--1438. {AAAI} Press.

\end{thebibliography}
\bibliographystyle{acl_natbib}

\appendix

\end{document}